\documentclass{article}

\usepackage{microtype}
\usepackage{graphicx}
\usepackage{booktabs} %

\usepackage{hyperref}

\usepackage[accepted]{icml2023}

\usepackage{amsmath}
\usepackage{amssymb}
\usepackage{mathtools}
\usepackage{amsthm}

\usepackage[capitalize,noabbrev]{cleveref}

\theoremstyle{plain}

\theoremstyle{definition}

\theoremstyle{remark}

\usepackage[textsize=tiny]{todonotes}

\usepackage{subcaption}
\usepackage{mathtools}
\usepackage{algpseudocode}
\usepackage{algorithm}
\usepackage{siunitx}
\usepackage[utf8]{inputenc}
\usepackage[T1]{fontenc}
\usepackage{etoolbox}

\robustify\bfseries
\newcommand{\std}[1]{{\fontsize{8}{9.6}\selectfont $\hspace{0.1em}\pm\hspace{0.1em}#1$}} %

\DeclarePairedDelimiter\abs{\lvert}{\rvert}

\newcommand{\distGaussian}[2]{\mathcal{N}(#1, #2)}
\newcommand{\distUniform}[2]{{U}(#1, #2)}
\newcommand{\rfSplitNodes}{N^{\text{split}}}
\newcommand{\rfLeafNodes}{N^{\text{leaf}}}

\icmltitlerunning{Neural Random Forest Imitation}

\begin{document}

\twocolumn[
\icmltitle{Neural Random Forest Imitation}

\begin{icmlauthorlist}
\icmlauthor{Christoph Reinders}{yyy}
\icmlauthor{Bodo Rosenhahn}{yyy}
\end{icmlauthorlist}

\icmlaffiliation{yyy}{Institute for Information Processing, L3S / Leibniz University Hannover, Germany}

\icmlcorrespondingauthor{Christoph Reinders}{reinders@tnt.uni-hannover.de}

\icmlkeywords{Imitation Learning, Neural Networks, Random Forests, Classification} 

\vskip 0.3in
]

\printAffiliationsAndNotice{}  %

\begin{abstract}
We present Neural Random Forest Imitation -- a novel approach for transforming random forests into neural networks. 
Existing methods propose a direct mapping and produce very inefficient architectures.
In this work, we introduce an imitation learning approach by generating training data from a random forest and learning a neural network that imitates its behavior.
This implicit transformation creates very efficient neural networks that learn the decision boundaries of a random forest.
The generated model is differentiable, can be used as a warm start for fine-tuning, and enables end-to-end optimization. 
Experiments on several real-world benchmark datasets demonstrate superior performance, especially when training with very few training examples. 
Compared to state-of-the-art methods, we significantly reduce the number of network parameters while achieving the same or even improved accuracy due to better generalization.
\end{abstract}

\section{Introduction}

Neural networks have become very popular in many areas, such as computer vision \citep{covmap_Krizhevsky2012AlexNet,covmap_chimeramix,covmap_renNIPS15fasterrcnn,covmap_Simonyan15,covmap_Zhao2017PSPNet,covmap_Qiao2021DetectoRSDO,covmap_RudWeh2022a,covmap_Sun2021FSCEFO}, speech recognition \citep{covmap_graves6638947,covmap_Park2019,covmap_Sun2021FSCEFO}, automated game-playing \citep{covmap_Mnih2015,covmap_DocDoe2017}, or natural language processing \citep{covmap_Collobert:2011:NLP:2078183.2078186,covmap_NIPS2014_a14ac55a,covmap_Otter2021ASO}. 
Researchers have published many datasets for training neural networks and put enormous effort into providing labels for each data sample. 
For real-world applications, the dependency on large amounts of labeled data represents a significant limitation \citep{covmap_reason:BreFriOlsSto84a,covmap_Hekler2019WhyWN,covmap_barz2020,covmap_Qi2020SmallDC,covmap_phoo2021STARTUP,covmap_WANG202161}. Frequently, there is little or even no labeled data for a particular task and hundreds or thousands of examples have to be collected and annotated.
This particularly affects new applications and rare labels (e.g., detecting rare diseases or defects in manufacturing).
Transfer learning and regularization methods are usually applied to reduce overfitting. 
However, for training with little data, the networks still have a considerable number of parameters that have to be fine-tuned -- even if just the last layers are trained.

In contrast to neural networks, random forests are very robust to overfitting due to their ensemble of multiple decision trees. Each decision tree is trained on randomly selected features and samples.
Random forests have demonstrated remarkable performance in many domains \citep{covmap_fernandez2014}.  
While the generated decision rules are simple and interpretable, the orthogonal separation of the feature space can also be disadvantageous on other datasets, especially with correlated features \citep{covmap_menze2011Oblique}.
Additionally, random forests are not differentiable and cannot be fine-tuned with gradient-based optimization.

The combination of neural networks and random forests brings both worlds together. 
Neural networks have demonstrated excellent performance in complex data modeling but require large amounts of training data.
Random forests are very good in learning with very little data without overfitting.
The advantages of our approach for mapping random forests into neural networks are threefold:
(1)~We enable the generation of neural networks with very few training examples. 
(2)~The resulting network can be used as a warm start, is fully differentiable, and allows further end-to-end fine-tuning. %
(3)~The generated network can be easily integrated into any trainable pipeline (e.g., jointly with feature extraction) and existing high-performance deep learning frameworks can be used directly. This accelerates the process and enables parallelization via GPUs.

Mapping random forests into neural networks is already used in many applications such as network initialization \citep{covmap_DBLP:journals/tnn/HumbirdPM19}, camera localization \citep{covmap_massiceti2017random}, object detection \citep{covmap_ReiAck2018a,covmap_ReiAck2019a}, or semantic segmentation \citep{covmap_BMVC2016_144}.
State-of-the-art methods %
\citep{covmap_massiceti2017random,covmap_Sethi1990RFNN,covmap_welbl2014casting} create a two-hidden-layer neural network by adding a neuron for each split node and each leaf node of the decision trees.
The number of parameters of the networks becomes enormous as the number of nodes grows exponentially with the increasing depth of the decision trees.
Additionally, many weights are set to zero so that an inefficient representation is created. Due to both reasons, the mappings do not scale and are only applicable to simple random forests.

In this work, we present an imitation learning approach to generate neural networks from random forests, which results in very efficient models. 
We introduce a method for generating training data from a random forest that creates any amount of input-target pairs. With this data, a neural network is trained to imitate the random forest.
Experiments demonstrate that the accuracy of the imitating neural network is equal to the original accuracy or even slightly better than the random forest due to better generalization while being significantly smaller. 
To summarize, our \textbf{contributions} are as follows:
\begin{itemize}
	\item We propose a novel method for implicitly transforming random forests into neural networks by generating data from a random forest and training an random forest-imitating neural network. Labeled data samples are created by evaluating the decision boundaries and guided routing to selected leaf nodes. 
	\item In contrast to direct mappings, our imitation learning approach is scalable to complex classifiers and deep random forests. 
	\item We enable learning and initialization of neural networks with very little data.
	\item Neural networks and random forests can be combined in a fully differentiable, end-to-end pipeline for acceleration and further fine-tuning.
\end{itemize}

\section{Related Work}

\begin{figure*}[t]
	\centering
	\includegraphics[width=1.0\textwidth]{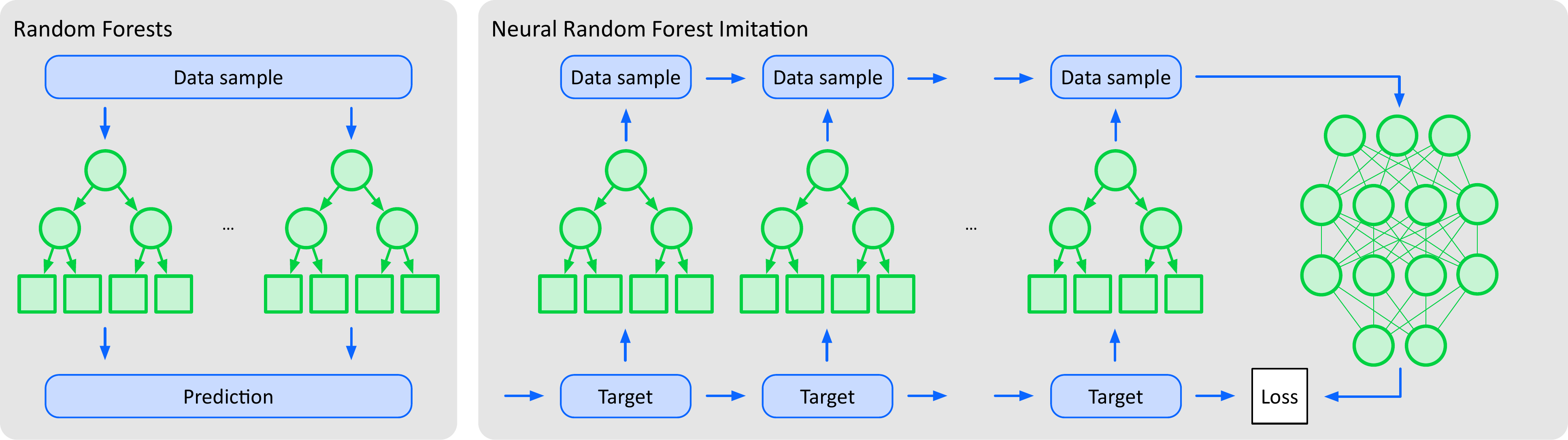}	
	\caption{
		Neural random forest imitation enables an implicit transformation of random forests into neural networks. Usually, data samples are propagated through the individual decision trees and the split decisions are evaluated during inference. 
		We propose a method for generating input-target pairs by reversing this process and training a neural network that imitates the random forest. 
		The resulting network is much smaller compared to current state-of-the-art methods, which directly map the random forest.
	} 
	\label{covmap_fig_overview}
\end{figure*}

Random forests and neural networks share some similar characteristics, such as the ability to learn arbitrary decision boundaries; however, both methods have different advantages. 
Random forests are based on decision trees. Various tree models have been presented -- the most well-known are C4.5 \citep{covmap_quinlan1993c45} and CART \citep{covmap_reason:BreFriOlsSto84a}. 
Decision trees learn rules by splitting the data. The rules are easy to interpret and additionally provide an importance score of the features.
Random forests \citep{covmap_breiman2001random} are an ensemble method consisting of multiple decision trees, with each decision tree being trained using a random subset of samples and features. 
\citet{covmap_fernandez2014} conduct extensive experiments comparing 179 classifiers on 121  UCI datasets \citep{covmap_Dua:2019}. The authors show that random forests perform best, followed by support vector machines with a radial basis function kernel. Therefore, random forests are often considered as a reference for new classifiers.

Neural networks are universal function approximators. 
The generalization performance has been widely studied. \citet{covmap_zhang2016understanding} demonstrate that deep neural networks are capable of fitting random labels and memorizing the training data. \citet{covmap_pmlr-v119-bornschein20a} analyze the performance across different dataset sizes. 
\citet{covmap_NEURIPS2018_fface838} evaluate the performance of modern neural networks using the same test strategy as \citet{covmap_fernandez2014} and find that neural networks achieve good results but are not as strong as random forests.

\citet{covmap_Sethi1990RFNN} presents a mapping of decision trees to two-hidden-layer neural networks. 
In the first hidden layer, the number of neurons equals the number of split nodes in the decision tree. Each of these neurons implements the decision function of the split nodes and determines the routing to the left or right child node.
The second hidden layer has a neuron per leaf node in the decision tree. Each of the neurons is connected to all split nodes on the path from the root node to the leaf node to evaluate if the data is routed to the respective leaf node. Finally, the output layer is connected to all leaf neurons and aggregates the results by implementing the leaf votes. 
By using hyperbolic tangent and sigmoid functions, respectively, as activation functions between the layers, the generated network is differentiable and, thus, trainable with gradient-based optimization algorithms. 
The method can be easily extended to random forests by mapping all trees.

\citet{covmap_welbl2014casting} and \citet{covmap_Biau2018} follow a similar strategy. The authors propose a method that maps random forests into neural networks as a smart initialization and then fine-tunes the networks by backpropagation. Two training modes are introduced: \textit{independent} and \textit{joint}. Independent training fits all networks one after the other and creates an ensemble of networks as a final classifier. Joint training concatenates all tree networks into one single network so that the output layer is connected to all leaf neurons in the second hidden layer from all decision trees and all parameters are optimized together. Additionally, the authors evaluate sparse and full connectivity. 
Sparse connectivity maintains the tree structures and has fewer weights to train. In practice, sparse weights require a special differentiable implementation, which can drastically decrease performance, especially when training on a GPU. Full connectivity optimizes all parameters of the fully connected network.
\citet{covmap_massiceti2017random} extend this approach and introduce a network splitting strategy by dividing each decision tree into multiple subtrees. The subtrees are mapped individually and share common neurons for evaluating the split decision.

These techniques, however, are only applicable to trees of limited depth. As the number of nodes grows exponentially with the increasing depth of the trees, inefficient representations are created, causing extremely high memory consumption.
In this work, we address this issue by proposing an imitation learning-based method that results in much more efficient models.

\section{Background and Notation}

In this section, we briefly describe decision trees \citep{covmap_reason:BreFriOlsSto84a}, random forests \citep{covmap_breiman2001random}, and the notation used throughout this work.
Decision trees consist of \textit{split nodes} $\rfSplitNodes$ and \textit{leaf nodes} $\rfLeafNodes$. Each split node $s \in \rfSplitNodes$ performs a \textit{split decision} and routes a data sample $x$ to the left or right child node, denoted as $\operatorname{c}_{\text{left}}(s)$ and $\operatorname{c}_{\text{right}}(s)$, respectively. When using binary, axis-aligned split decisions, a single feature $f(s) \in \{1, \dots, N\}$ and a threshold $\theta(s) \in \mathbb{R}$ are the basis for the split, where $N$ is the number of features. 
If the value of feature $f(s)$ is smaller than $\theta(s)$, the data sample is routed to the left child node and otherwise to the right child node, denoted as
\begin{align}
	x \in \operatorname{c}_{\text{left}}(s) &\iff x_{\operatorname{f}(s)} < \theta(s) \\
	x \in \operatorname{c}_{\text{right}}(s) &\iff x_{\operatorname{f}(s)} \geq \theta(s).
\end{align}

Data samples are routed through a decision tree until a leaf node $l \in \rfLeafNodes$ is reached which stores the target value. For the classification task, these are the estimated class probabilities $P_{\text{leaf}}(l) = (p^{l}_1, \dots, p^{l}_C)$, where $C$ is the number of classes. 
Decision trees are trained by creating a root node and consecutively finding the best split of the data based on a criterion. The resulting subsets are assigned to the left and right child node, and the subtrees are processed further. Commonly used criteria are the \textit{Gini Impurity} or \textit{Entropy}.

A single decision tree is very fast and operates on high-dimensional data. However, it tends to overfit the training data by constructing a deep tree that separates perfectly all training examples. While having a very small training error, this easily results in a large test error.
Random forests address this problem by learning an ensemble of $n_T$ decision trees.
Each tree is trained with a random subset of training examples and features. The prediction $ \operatorname{RF}(x)$ of a random forest is calculated by averaging the predictions of all decision trees.

\section{Neural Random Forest Imitation}

Our proposed approach, called \textit{Neural Random Forest Imitation} (NRFI), implicitly transforms random forests into neural networks. 
The main concept includes (1) generating training data from decision trees and random forests, (2)  adding strategies for reducing conflicts and increasing the variety of the generated examples, and (3) training a neural network that imitates the random forest by learning the decision boundaries.
As a result, NRFI enables the transformation of random forests into efficient neural networks. An overview of the proposed method is shown in Figure~\ref{covmap_fig_overview}.

\subsection{Data Generation}
\label{sec_data_generation}

First, we propose a method for generating data from a given random forest. 
A data sample $x \in \mathbb{R}^N$ is an $N$-dimensional vector, where $N$ is the number of features. We select a target class $t \in [1, \dots, C]$ from $C$ classes and generate a data sample for the selected class.

\subsubsection{Data Initialization}

A data sample $x$ is initialized randomly. 
In the following, the feature-wise minimum and maximum of the training samples will be denoted as  $f_{\text{min}}, f_{\text{max}} \in \mathbb{R}^N$.
To initialize $x$, we sample $x \sim \distUniform{f_{\text{min}}}{f_{\text{max}}}$. In the next step, we will present a method for adapting the data sample to obtain characteristics of the target class.

\subsubsection{Data Generation from Decision Trees}
\label{covmap_sec_data_generation_from_tree}

\begin{figure*}[t]
    \centering
    \begin{subfigure}[]{0.49\linewidth}
        \includegraphics[width=\linewidth]{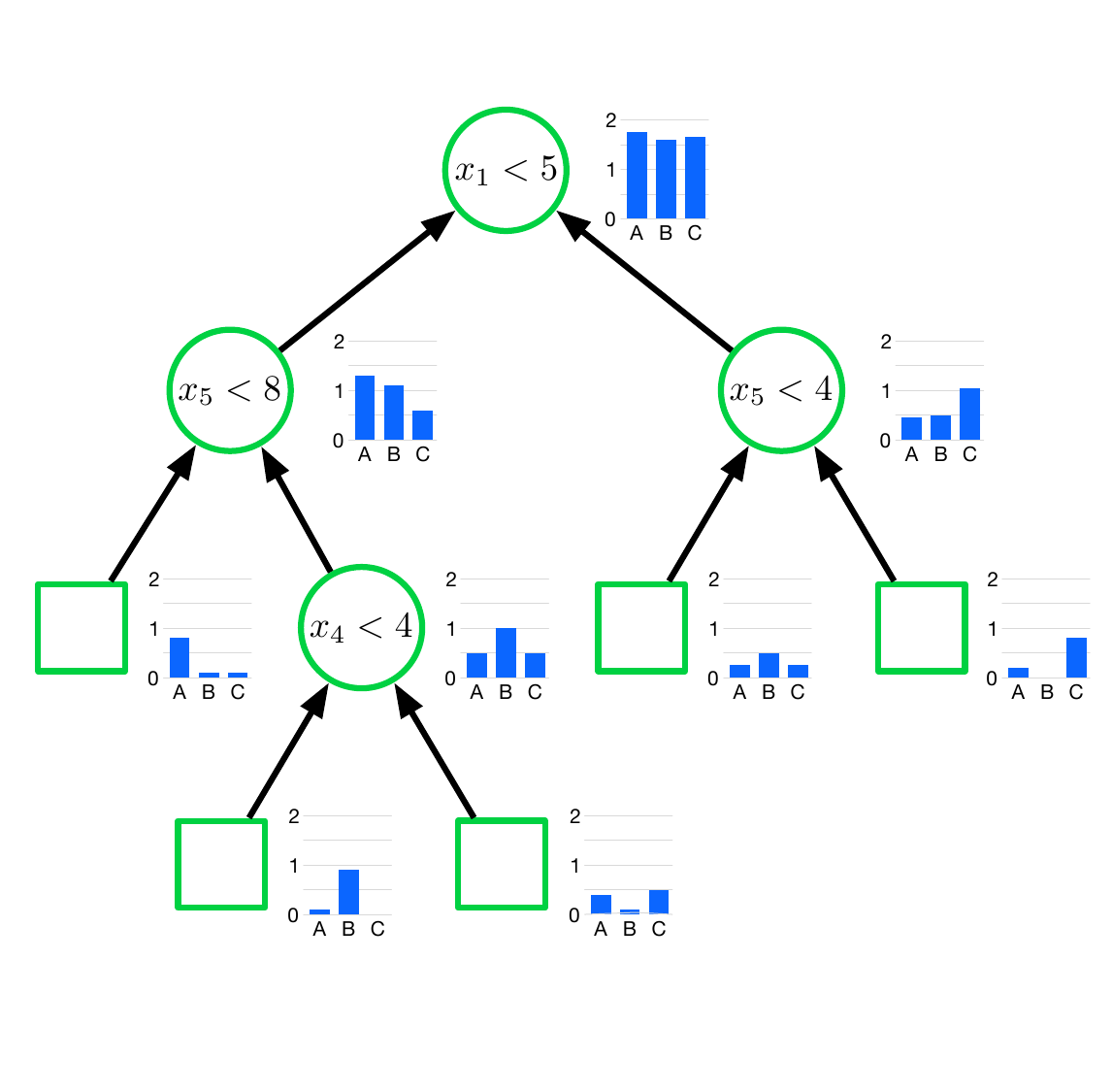}
        \caption{}
        \label{covmap_fig_data_generation_tree_left}
    \end{subfigure}
    \begin{subfigure}[]{0.49\linewidth}
        \includegraphics[width=\linewidth]{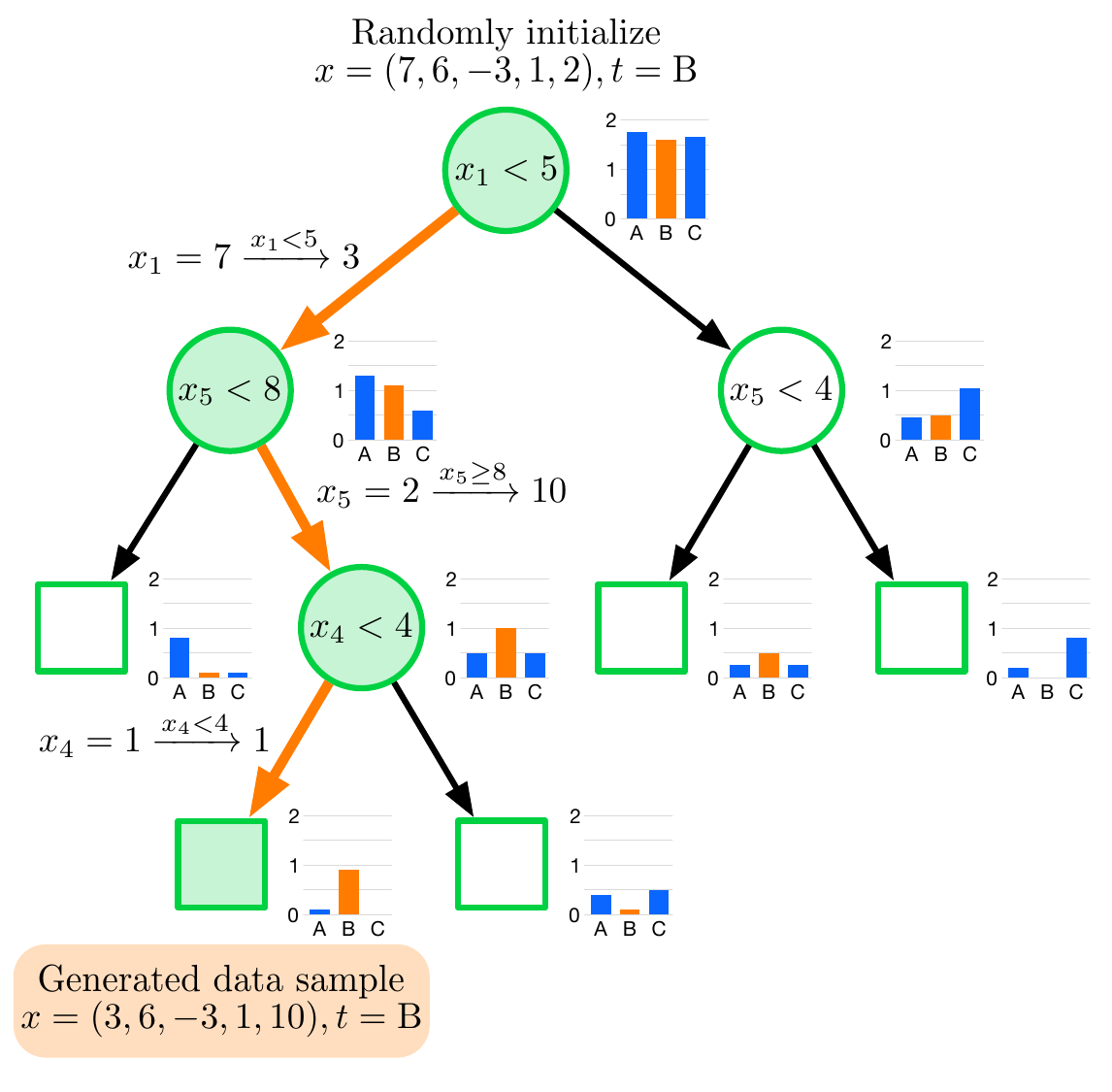}
        \caption{}
        \label{covmap_fig_data_generation_tree_right}
    \end{subfigure}
	\caption{Overview of the data generation process from a decision tree. First, the class distribution information is propagated from the leaf nodes to the split nodes (a). Afterward, data samples are generated by guided routing (Section \ref{covmap_sec_data_generation_from_tree}) and modifying the data based on the split decisions (b).
		The weights for sampling the left or right child node are highlighted in orange.}
	\label{covmap_fig_data_generation_tree}
\end{figure*}

A decision tree processes an input vector $x$ by routing the data through the tree until a leaf is reached. At each node, a split decision is evaluated, and the input is passed to the left child node or the right child node. Finally, a leaf $l$ is reached which stores the estimated probabilities $P_{\text{leaf}}(l) = (p^{l}_1, \dots, p^{l}_C)$ for each class. 

We reverse this process and present a method for generating training data from a decision tree. 
An overview of the proposed data generation process is shown in Figure~\ref{covmap_fig_data_generation_tree}.
First, the class distribution information is propagated bottom-up from the leaf nodes to the split nodes (see Figure~\ref{covmap_fig_data_generation_tree_left}) and we define the class weights $W(n) = (w^n_1, \dots, w^n_C)$ for every node $n$ as follows: 
\begin{equation}
	W(n) = \begin{dcases}
		P_{\text{leaf}}(n) &\text{if}  \enspace n \in \rfLeafNodes \\
		W(\operatorname{c}_{\text{left}}(n)) + W(\operatorname{c}_{\text{right}}(n)) & \text{if}  \enspace n \in \rfSplitNodes
	\end{dcases}
\end{equation}
For every leaf node, the class weights are equal to the stored probabilities in the leaf. For every split node, the class weights in the child nodes are summed up. 

After preparation, data samples for a target class $t$ are generated (see Figure~\ref{covmap_fig_data_generation_tree_right}). 
For that, characteristics of the target class are successively added to the data sample. Starting at the root node, we modify the input data so that it is routed through selected split nodes until a leaf node is reached. 
The pseudocode is presented in Algorithm~\ref{covmap_algo_compression_generate_data_tree}. 

The routing is guided based on the weights for the target class in the left child node $w_{\text{left}} =  w^{\operatorname{c}_{\text{left}}(n)}_t$ and right child node $w_{\text{right}} = w^{\operatorname{c}_{\text{right}}(n)}_t$. 
The weights are normalized by their L2-norm, denoted as  $\hat{w}_{\text{left}}$ and  $\hat{w}_{\text{right}}$.
Afterward, the left or right child node is randomly selected as next child node $n_{\text{next}}$ depending on $\hat{w}_{\text{left}}$ and $\hat{w}_{\text{right}}$.

In the next step, the data sample is updated. We verify that the data sample is routed to the selected child node by evaluating the split decision. A split node $s$ routes the data to the left or right child node based on a split feature $\operatorname{f}(s)$ and a threshold $\theta(s)$. If the value of the split feature  $x_{\operatorname{f}(s)}$ is smaller than $\theta(s)$, the data sample is routed to the left child node and otherwise to the right child node.
The data sample is modified if it is not already routed to the selected child node by assigning a new value. If the selected child node is the left child node, the value has to be smaller than the threshold $\theta(s)$ and a new value within the minimum feature value $f_{\text{min}, \operatorname{f}(s)}$ and $\theta(s)$  is randomly sampled:
\begin{align}
	x_{\operatorname{f}(s)} \sim \distUniform{f_{\text{min}, \operatorname{f}(s)}}{\theta(s)}.
\end{align}
If the data sample is supposed to be routed to the right child node, the new value is randomly sampled between $\theta(s)$ and the maximum feature value $f_{\text{max}, \operatorname{f}(s)}$:
\begin{align}
	x_{\operatorname{f}(s)} \sim \distUniform{\theta(s)}{f_{\text{max}, \operatorname{f}(s)}}.
\end{align}
This process is repeated until a leaf node is reached. In each node, characteristics are added that classify the data sample as the target class.

During this process, modifications can conflict with previous decisions because features are used multiple times within a decision tree or across multiple decision trees. 
Therefore, the current routing is weighted with a factor $w_{\text{path}} \geq 1$ to prioritize the path and not change the data sample if possible. 
Overall, the presented method enables the generation of data samples and corresponding labels from a decision tree without adding any further data.

\begin{algorithm}[t]
	\caption{\textsc{DataGenerationFromTree} \newline Generate data samples from a decision tree}
	\label{covmap_algo_compression_generate_data_tree}
	\textbf{Input:} Decision tree split features $\operatorname{f}(n)$ and thresholds  $\theta(n)$, target class $t$, feature minimum $f_{\text{min}}$ and maximum $f_{\text{max}}$, 
	class weights $W(n) = (w^n_1, \dots, w^n_C)$ for all nodes $n \in \rfSplitNodes \cup \rfLeafNodes $ \\
	\textbf{Output:} {Data sample for target class $t$}
	\begin{algorithmic}[1]
		\State Sample $x \sim \distUniform{f_{\text{min}}}{f_{\text{max}}} \in \mathbb{R}^N$
		\State $n \gets $ root node		
		\While{$n \not\in \rfLeafNodes $}
		\State $w_{\text{left}} \gets w^{\operatorname{c}_{\text{left}}(n)}_t$
		\State $w_{\text{right}} \gets w^{\operatorname{c}_{\text{right}}(n)}_t$
		\If{feature $\operatorname{f}(n)$ is already used}
		\State weight current route with $w_{\text{path}}$
		\State $w_{\text{current}} \gets w_{\text{current}} \cdot w_{\text{path}} $
		\EndIf
		\State $\hat{w}_{\text{left}}, \hat{w}_{\text{right}} \gets $ normalize $w_{\text{left}}$ and $w_{\text{right}}$ 
		\State $n_{\text{next}} \gets$ randomly select left or right child node with probability of $\hat{w}_{\text{left}}$ and $\hat{w}_{\text{right}}$, respectively 
		\If{$n_{\text{next}} = \operatorname{c}_{\text{left}}(n)$} \label{covmap_algo_data_tree_line_left_cl}
		\If{$x_{\operatorname{f}(n)}  \geq \theta(n)$} \label{algo_data_tree_line_left_x}
		\State $x_{\operatorname{f}(n)} \sim \distUniform{f_{\text{min}, \operatorname{f}(n)}}{\theta(n)}$ \label{covmap_algo_data_tree_line_left_random}
		\EndIf				
		\Else \label{algo_data_tree_line_right_rl}
		\If{$x_{\operatorname{f}(n)}  < \theta(n)$} \label{algo_data_tree_line_right_x}
		\State $x_{\operatorname{f}(n)} \sim  \distUniform{\theta(n)}{f_{\text{max}, \operatorname{f}(n)}}$ \label{covmap_algo_data_tree_line_right_random}
		\EndIf		
		\EndIf
		\State mark feature $\operatorname{f}(n)$ as used
		\State $n \gets n_{\text{next}}$
		\EndWhile
		\State \Return $x$
	\end{algorithmic}
\end{algorithm}

\subsubsection{Data Generation from Random Forests}

In the next step, we extend the method to generate data from random forests.
Random forests consist of $n_T$ decision trees $\operatorname{RF} = \{T_1, \dots, T_{n_T}\}$.
For generating a data sample $x$, the presented method for a single decision tree is applied to multiple decision trees consecutively.
The initialization is performed only once and the visited features are shared.
In each decision tree, the data sample is modified and routed to selected nodes based on the target class $t$.
When using all decision trees, data samples are created where all trees agree with a high probability.
For generating examples with varying confidence, i.e., the predictions of the individual decision trees diverge, we select a subset of $n_{\text{sub}}$ decision trees $\operatorname{RF}_{\text{sub}}\subseteq \operatorname{RF} $.

All decision trees in $\operatorname{RF}_{\text{sub}}$ are processed in random order to generate a data sample. For each decision tree, the presented method modifies the data sample based on the target class. 
Finally, the output of the random forest $y = \operatorname{RF}(x)$ is predicted. 
In most cases, the prediction matches the target class. 
Due to factors such as the stochastic process, a small subset size, or varying predictions of the decision trees, it can be different occasionally.
Thus, an input-target pair $(x, y)$ has been created, showing similar characteristics as the target class and any amount of data can be generated by repeating this process. 

\subsubsection{Automatic Confidence Distribution}
\label{covmap_sec_confidence_distribution}

The number of decision trees $n_{\text{sub}}$ can be set to a fixed value or sampled uniformly. Alternatively, we present an automatic process for determining an optimal distribution of the confidences for generating a wide variety of different examples. The strategy is motivated by \textit{importance weighting} \citep{covmap_DBLP:conf/nips/FangL0S20}.
We generate $n$ data samples ($n$ is empirically set to $1000$) for each number of decision trees $j \in [1, n_T]$.
The respective generated datasets will be denoted as $D_{j}$. 

An optimal sampling process generates highly diverse data samples with different confidences.
To achieve that, an automated balancing of the distributions is determined. A histogram with $H$ bins is calculated for each $D_{j}$, where $h^{j}_{i}$ denotes the number of generated examples in the $i$-th interval (equally distributed) from the distribution with $j$ decision trees. In the next step, a weight $w^{D}_j$ is defined for each distribution, and we optimize $w^{D}$ as follows:
\begin{multline}
	\min_{w^{D}}  \left\lVert \left[\sum_{j=1}^{n_T} w^{D}_{j} h^{j}_{1} \ \dots \ \sum_{j=1}^{n_T} w^{D}_{j} h^{j}_{H}\right]^{T} -  \begin{bmatrix} 1 \\ \vdots \\ 1 \end{bmatrix} 
	\right\rVert^{2} \\ \quad \textrm{s.t.} \quad \forall_{j} \ 0 \leq w^{D}_{j},
\end{multline}
where $w^{D} \in \mathbb{R}^{n_T}$. This optimization finds a weighting of the number of decision trees so that the generated confidences cover the full range equally. For that, the number of samples per bin $h^{j}_{i}$ is summed up, weighted over all numbers of decision trees. After determining $w^{D}$, the number of decision trees can be sampled depending on $w^{D}_j$. 
An analysis of different sampling methods will be presented in Section~\ref{covmap_sec_ablation_study_sampling}.
Automatically balancing the number of decision trees generates data samples with low and high confidence very equally distributed.
The process does not require training data and provides a universal solution.

\subsection{Imitation Learning}

Finally, a neural network that imitates the random forest is trained. The network learns the decision boundaries from the generated data and approximates the same function as the random forest.
The network architecture is based on a fully connected network with one or multiple hidden layers. 
The data dimensions are the same as those of the random forest, i.e., an $N$-dimensional input and $C$-dimensional output.
Each hidden layer is followed by a ReLU activation~\citep{covmap_Nair:2010:ReLU}. The last fully connected layer is using a softmax activation.

For training, we generate input-target pairs $(x, y)$ as described in the last section. 
These training examples are fed into the training process to teach the network to predict the same results as the random forest. To avoid overfitting, the data is generated on-the-fly so that each training example is unique. In this way, we learn an efficient representation of the decision boundaries and are able to transform random forests into neural networks implicitly. 
In addition to that, the training is performed end-to-end on the generated data, and we can easily integrate the original training data.

\section{Experiments}

In this section, we perform several experiments to analyze the performance of neural random forest imitation and compare our method to state-of-the-art methods. 
In the following, we evaluate on standard benchmark datasets to present a general approach for various domains. While we focus on classification tasks in this work, NRFI can be simply adapted for regression tasks.

\subsection{Datasets}

The experiments are evaluated on nine classification datasets from the UCI Machine
Learning Repository \citep{covmap_Dua:2019} (\textit{Car}, \textit{Connect-4}, \textit{Covertype}, \textit{German Credit}, \textit{Haberman}, \textit{Image Segmentation}, \textit{Iris}, \textit{Soybean}, and \textit{Wisconsin Breast Cancer (Original)}). The datasets cover many real-world problems in different areas, such as finance, computer vision, games, or medicine. %

Following \citet{covmap_fernandez2014}, each dataset is split into a training and a test set using a 50/50 split while maintaining the label distribution. Afterward, the number of training examples is limited to $n_{\text{limit}}$ examples per class. We evaluate the training with $5$, $10$, $20$, and $50$ examples per class.
In contrast to \citet{covmap_fernandez2014}, we extract validation sets from the training set (e.g., for hyperparameter tuning). This ensures that the training and validation data are not mixed with the test data. For some datasets which provide a separate test set, the test accuracy is evaluated on the respective set.
Missing values are set to the mean value of the feature.
All experiments are repeated ten times with randomly sampled splits. The methods are repeated additionally four times with different seeds on each split.

\subsection{Implementation Details}
In all our experiments, stochastic gradient descent with Nesterov momentum as optimizer and cross-entropy loss are used. 
The initial learning rate is set to $0.1$, momentum to $0.9$, and weight decay to $0.0005$. The batch size is set to $128$ and $512$, respectively, for generated data.
The input data is normalized to $[-1, 1]$.
For generating a wide variety of data, the prioritization of the current path $w_{\text{path}} \sim 1 + \abs{\distGaussian{0}{5}}$ is sampled for each data sample individually.
A new random forest is trained every $100$ epochs to average the influence of the stochastic process, and the generated data samples are mixed.
In the following, training on generated data will be denoted as \textit{NRFI (gen)} and training on generated and original data as \textit{NRFI (gen+ori)}. The fraction of NRFI data is set to $0.9$.
Random forests are trained with $500$ decision trees, which are commonly used in practice \citep{covmap_fernandez2014,covmap_NEURIPS2018_fface838}. 
The decision trees are constructed up to a maximum depth of ten. For splitting, the Gini Impurity is used and $\sqrt{N}$ features are randomly selected, where $N$ is the number of features.

\subsection{Results}

\begin{figure*}[t] %
	\centering
	\includegraphics[width=1.0\textwidth]{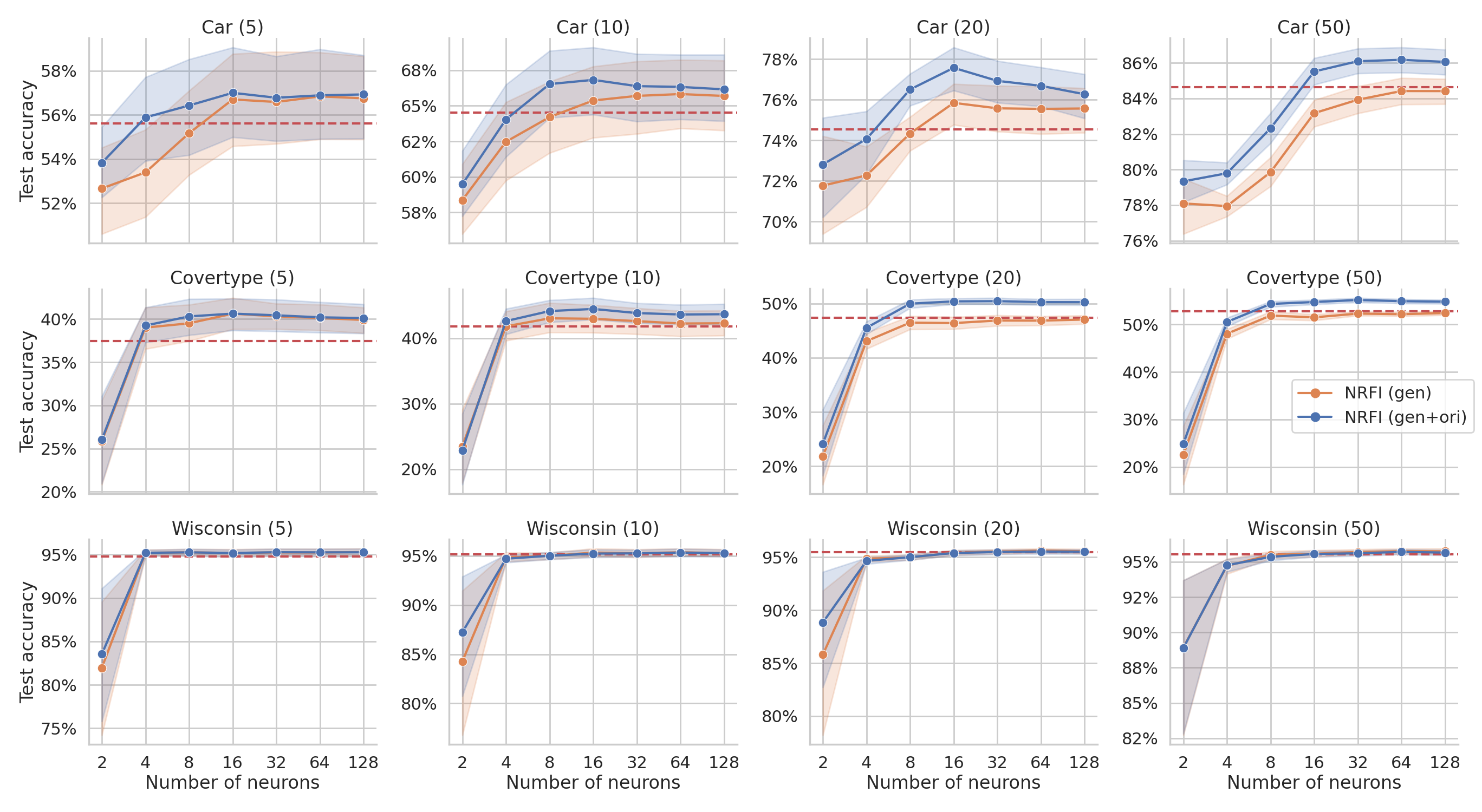}
	\caption{
		Test accuracy depending on the network architecture (i.e., number of neurons in both hidden layers). 
		Different datasets are shown per row, with an increasing number of training examples per class from left to right (indicated in parentheses). 
		The red dashed line shows the accuracy of the random forest. 
		NRFI with generated data is shown in orange and NRFI with generated and original data in blue.
		With increasing network capacity, NRFI is capable of imitating and even outperforming the random forest.
	} 
	\label{covmap_fig_experiment_accuracy}
\end{figure*}

The proposed method generates data from a random forest and trains a neural network that imitates the random forest. The goal is that the neural network approximates the same function as the random forest. This also implies that the network reaches the same accuracy if successful. 

We analyze the performance by training random forests for each dataset and evaluating neural random forest imitation with different network architectures. 
A variety of network architectures with different depths, widths, and additional layers such as Dropout have been studied. In this work, we focus on two-hidden-layer networks with an equal number of neurons in both layers for clarity. 
The results are shown in Figure~\ref{covmap_fig_experiment_accuracy} exemplarily for the \textit{Car}, \textit{Covertype}, and \textit{Wisconsin Breast Cancer (Original)} dataset. The other datasets show similar characteristics. The overall evaluation on all datasets is presented in the next section. 
The number of training examples per class is shown in parentheses and increases in each row from left to right.
For each setting, the test accuracy of the random forest is indicated by a red dashed line.
The average test accuracy and standard deviation depending on the network architecture, i.e., the number of neurons in the first and second hidden layer, are plotted for different architectures.
NRFI (gen), which is trained with generated data only, is shown in orange and NRFI (gen+ori), which is trained with generated and original data, is shown in blue.

The analysis shows that the accuracy of the neural networks trained by NRFI reaches the accuracy of the random forest for all datasets. Only very small networks do not have the required capacity.
The proposed method for generating labeled data from random forests by analyzing the decision boundaries enables training neural networks that imitate the random forests. 
For instance, in the case of $5$ training examples per class, a two-hidden-layer network with $16$ neurons in both layers already achieves the same accuracy as the random forest across all three datasets in Figure~\ref{covmap_fig_experiment_accuracy}. 
Additionally, the experiment shows that the training is very robust to overfitting even when the number of parameters in the network increases. 
When combining the generated data and original data, the accuracy on \textit{Car} and \textit{Covertype} improves with an increasing number of training examples.

Overall, the experiment shows that the accuracy increases with an increasing number of neurons in both layers and NRFI is robust to different network architectures. 
NRFI is capable of generating a large variety of unique examples from random forests which have been initially trained on a limited amount of data.

\begin{figure*}[t] %
	\centering
	\includegraphics[width=0.9\textwidth]{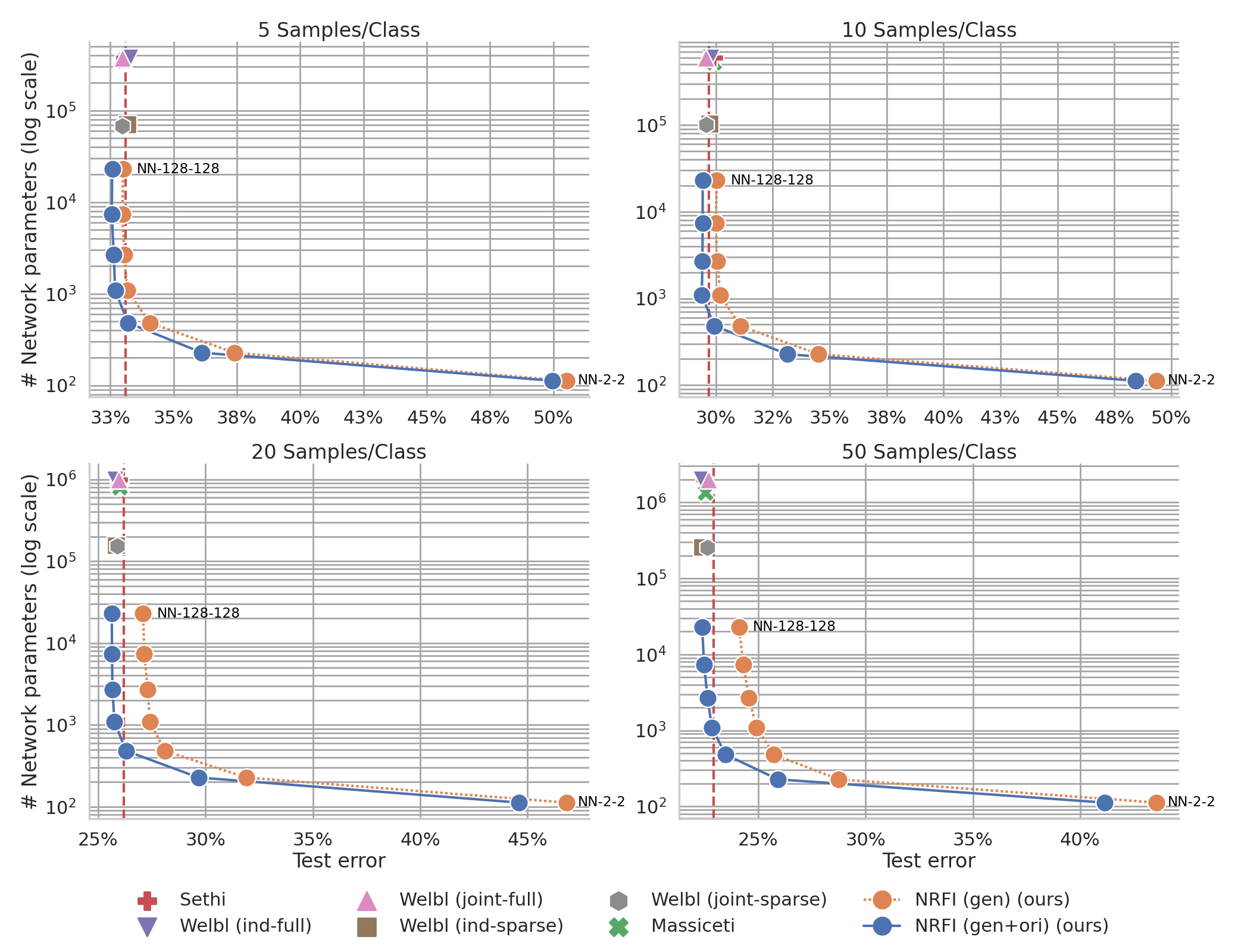}	
	
	\caption{
		Comparison of the state-of-the-art and our proposed method for transforming random forests into neural networks.
		The closer a method is to the lower-left corner, the better it is (fewer number of network parameters and lower test error).
		For neural random forest imitation, different network architectures are shown.
		Note that the number of network parameters is shown on a logarithmic scale.}
	\label{covmap_fig_compare_network_size_vs_error}
\end{figure*}

\begin{table*}[t] %
	\centering
	\setlength{\tabcolsep}{2.5pt} %
	
	\begin{tabular}{l
			S[table-format =2.2, table-space-text-post=\std{9.99},separate-uncertainty=true]
			S[table-format =2.2, table-space-text-post=\std{9.99},separate-uncertainty=true]
			S[table-format =2.2, table-space-text-post=\std{9.99},separate-uncertainty=true]
			S[table-format =2.2, table-space-text-post=\std{9.99},separate-uncertainty=true]|
			S[table-format =2.2, table-space-text-post=\std{9.99},separate-uncertainty=true]}
		\toprule
		& \multicolumn{4}{c}{Samples per class}& \\
		Method & {5} & {10} & {20} & {50} & {mean} \\
		\midrule
		
		DT                    &  62.95 \std{5.41} &  66.89 \std{4.18} &  70.82 \std{2.93} &  73.66 \std{2.20} &  68.58 \std{3.68} \\
		SVM                   &  65.21 \std{4.81} &  68.15 \std{4.44} &  71.91 \std{3.33} &  75.96 \std{2.22} &  70.31 \std{3.70} \\
		RF                    &  66.91 \std{4.01} &  70.31 \std{3.86} &  73.81 \std{2.46} &  77.08 \std{1.90} &  72.03 \std{3.06} \\
		NN                    &  65.50 \std{5.15} &  69.89 \std{4.13} &  73.11 \std{3.19} &  76.50 \std{2.53} &  71.25 \std{3.75} \\
		Sethi                 &  66.93 \std{4.01} &  70.06 \std{4.28} &  74.00 \std{3.00} &  77.50 \std{2.23} &  72.12 \std{3.38} \\
		Welbl (ind-full)      &  66.72 \std{4.04} &  70.21 \std{3.91} &  74.19 \std{2.50} &  \bfseries 77.63 \std{1.81} &  72.19 \std{3.06} \\
		Welbl (joint-full)    &  67.01 \std{4.14} &  70.42 \std{4.07} &  74.02 \std{2.80} &  77.31 \std{1.76} &  72.19 \std{3.19} \\
		Welbl (ind-sparse)    &  66.81 \std{4.07} &  70.27 \std{4.15} &  74.14 \std{2.58} &  77.60 \std{1.82} &  72.20 \std{3.15} \\
		Welbl (joint-sparse)  &  67.02 \std{4.17} &  70.41 \std{4.11} &  74.09 \std{2.77} &  77.36 \std{1.61} &  72.22 \std{3.17} \\
		Massiceti             &  66.97 \std{4.05} &  70.07 \std{4.28} &  73.98 \std{3.05} &  77.45 \std{2.26} &  72.12 \std{3.41} \\
 		\midrule
		NRFI (gen) (ours)     &  66.99 \std{4.09} &  69.95 \std{4.21} &  72.90 \std{2.67} &  75.90 \std{2.22} &  71.44 \std{3.30} \\
		NRFI (gen+ori) (ours) &  \bfseries 67.42 \std{4.15} &  \bfseries 70.57 \std{4.05} &  \bfseries 74.36 \std{2.44} &  77.62 \std{1.90} & \bfseries  72.49 \std{3.14} \\
		
		\bottomrule
	\end{tabular}
	\caption{
		Average test accuracy [\%] and standard deviation on all nine datasets for different numbers of training examples per class. The overall performance of each method is summarized in the last column. The best methods are highlighted in bold.
	}
	\label{covmap_table_accuracy}
\end{table*}

\subsection{Comparison to State of the Art}

We now compare the proposed method to state-of-the-art methods for mapping random forests into neural networks and classical machine learning classifiers such as random forests and support vector machines with a radial basis function kernel that have shown to be the best two classifiers across all UCI datasets \citep{covmap_fernandez2014}. In detail, we will evaluate the following methods:
\begin{itemize}
	\item DT: A decision tree \citep{covmap_reason:BreFriOlsSto84a} learns simple and interpretable split decisions to classify data. The Gini Impurity is used for splitting.
	\item SVM: Support vector machine \citep{covmap_libsvm} is a popular classifier that tries to find the best hyperplane that maximizes the margin between the classes. %
	As evaluated by \citet{covmap_fernandez2014}, the best performance is achieved with a radial basis function kernel.
	\item RF: Random forest \citep{covmap_breiman2001random} is an ensemble-based method consisting of multiple decision trees. Each decision tree is trained on a different randomly selected subset of features and samples. The classifier follows the same overall setup, i.e., $500$ decision trees and a maximum depth of ten.
	\item NN: A neural network \citep{covmap_hinton1988} with two hidden layers is trained using ReLU activation and cross-entropy loss. Possible values for the initial learning rate are $\{0.1, 0.01, 0.001, 0.0001, 0.00001\}$ and $\{2, 4, 8, 16, 32, 64, 128\}$ for the number of neurons in both hidden layers. The best hyperparameters are selected by performing a 4-fold cross-validation. 
	\item Sethi: The method proposed by \citet{covmap_Sethi1990RFNN} maps a random forest into a two-hidden-layer neural network by adding a neuron for each split node and each leaf node. The weights are set corresponding to the split decisions.
	\item Welbl: \citet{covmap_welbl2014casting} and \citet{covmap_Biau2018} present a similar mapping with subsequent fine-tuning. The authors introduce two training modes: \textit{independent} and \textit{joint}. The first optimizes each small network individually, while the latter joins all mapped decision trees into one network. Additionally, the authors evaluate a network with sparse connections and regular fully connected networks  (denoted as \textit{sparse} and \textit{full}).
	\item Massiceti: \citet{covmap_massiceti2017random} present a network splitting strategy to reduce the number of network parameters. The decision trees are divided into subtrees and mapped individually while sharing common split nodes. The optimal depth of the subtrees is determined by evaluating all possible values. 
\end{itemize}

First, we analyze the performance of state-of-the-art methods for mapping random forests into neural networks and neural random forest imitation. The results are shown in Figure~\ref{covmap_fig_compare_network_size_vs_error} for different numbers of training examples per class.
For each method, the average number of parameters of the generated networks across all datasets is plotted depending on the test error. That means that the methods aim for the lower-left corner (smaller number of network parameters and higher accuracy). Please note that the y-axis is shown on a logarithmic scale. 
The average performance of the random forests is indicated by a red dashed line.

The analysis shows that  Sethi, Welbl (ind-full), and Welbl (joint-full) generate the largest networks. 
Network splitting \citep{covmap_massiceti2017random} slightly improves the number of parameters of the networks. 
Using a sparse network architecture reduces the number of parameters. However, it should be noted that this requires special operations.
NRFI with and without the original data is shown for different network architectures. The smallest architecture has $2$ neurons in both hidden layers and the largest $128$. For NRFI (gen-ori), we can see that a network with $16$ neurons in both hidden layers (NN-16-16) is already sufficient to learn the decision boundaries of the random forest and achieve the same accuracy. When fewer training samples are available, NN-8-8 already has the required capacity. 
In the following, we will further analyze the accuracy and number of network parameters.

\subsubsection{Accuracy}

The average test accuracy and standard deviation for all methods are shown in Table~\ref{covmap_table_accuracy}. 
Here, we additionally include decision trees, support vector machines, random forests, and neural networks in the comparison. The evaluation is performed on all nine datasets, and results for different numbers of training examples are shown (increasing from left to right). The overall performance of each method is summarized in the last column.
For neural random forest imitation, a network architecture with $128$ neurons in both hidden layers is used. From the analysis, we can make the following observations: 
(1) When training neural random forest imitation with generated data only, the method achieves $99.18\%$ of the random forest accuracy ($71.44\%$ compared to $72.03\%$).
This shows that NRFI is capable of learning the decision boundaries. 
(2) Overall, NRFI trained with generated and original data reaches state-of-the-art performance ($50$ samples per class) or outperforms the other methods ($5$, $10$, and $20$ samples per class).

\subsubsection{Network Parameters}

\begin{table*}[t] %
	\centering
	\setlength{\tabcolsep}{6pt}	
	
	\begin{tabular}{
			lS[table-format = 6.0]
			S[table-format = 6.0]
            S[table-format = 6.0]
            S[table-format = 7.0]|
            S[table-format = 6.0]}
		\toprule
		& \multicolumn{4}{c}{Samples per class} & \\
		Method & {5} & {10} & {20} & {50} & {mean} \\
		\midrule
		& \multicolumn{5}{c}{Number of network parameters} \\
		Sethi                & 374299 & 592384 & 985294 & 1973341 & 981330 \\
		Welbl (ind-full)     & 374729 & 592147 & 984626 & 1972604 & 981027 \\
		Welbl (joint-full)   & 371965 & 589220 & 981816 & 1968118 & 977780 \\
		Welbl (ind-sparse)   &  70070 & 102895 & 154740 &  254344 & 145512 \\
		Welbl (joint-sparse) &  67344 & 100131 & 151944 &  251598 & 142754 \\
		Massiceti            & 348972 & 522640 & 792410 & 1328731 & 748188 \\
		NRFI (ours)      &   \bfseries 2676 &   \bfseries 2676 &   \bfseries 2676 &    \bfseries 2676 &   \bfseries 2676 \\
		
		\bottomrule
	\end{tabular}
	
	\caption{
		Comparison to state-of-the-art methods. For each method, the average number of parameters of the generated neural networks is shown.
		While achieving the same or even slightly better accuracy, neural random forest imitation generates much smaller models, enabling the mapping of complex random forests.
	}
	\label{covmap_table_network_size}
\end{table*}

\begin{figure*}[t] %
	\centering
    \begin{subfigure}{.8\linewidth}
        \includegraphics[width=\linewidth]{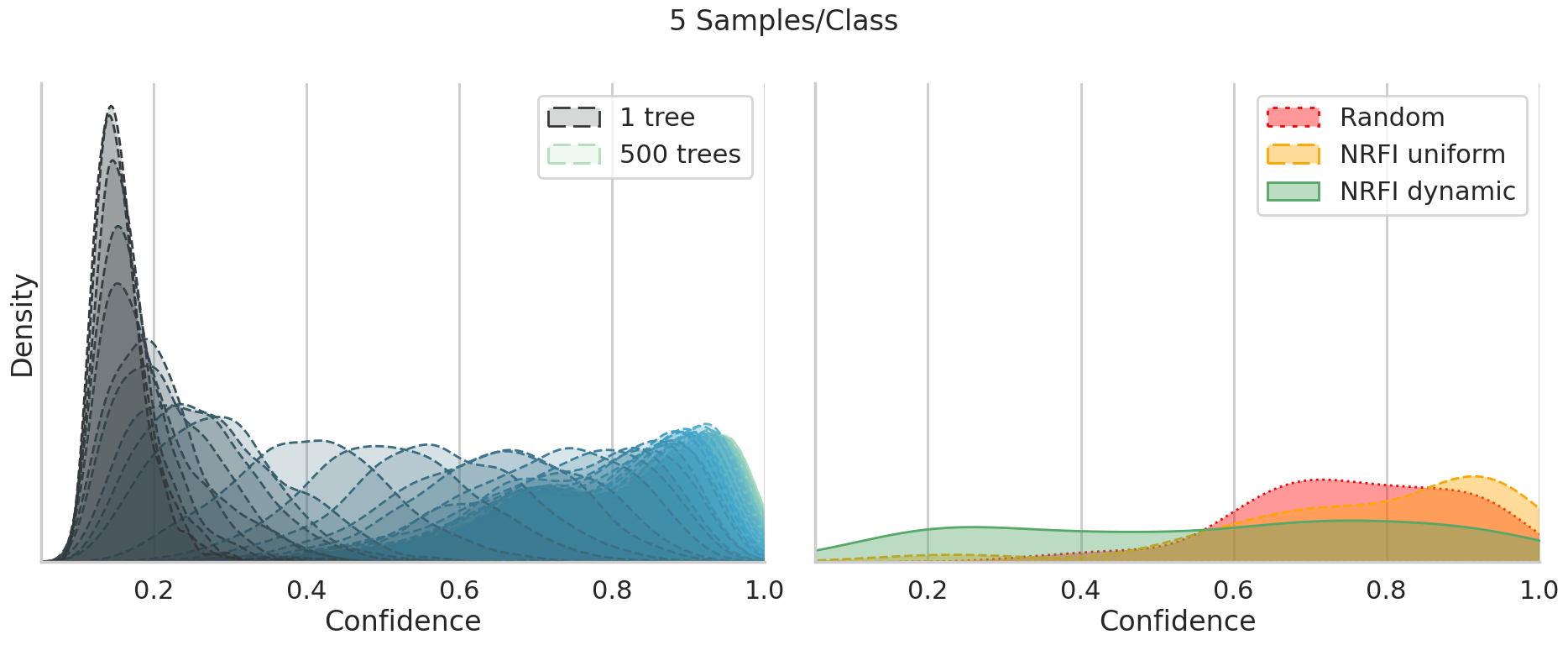}
    \end{subfigure}\\
    \begin{subfigure}{.8\linewidth}
        \includegraphics[width=\linewidth]{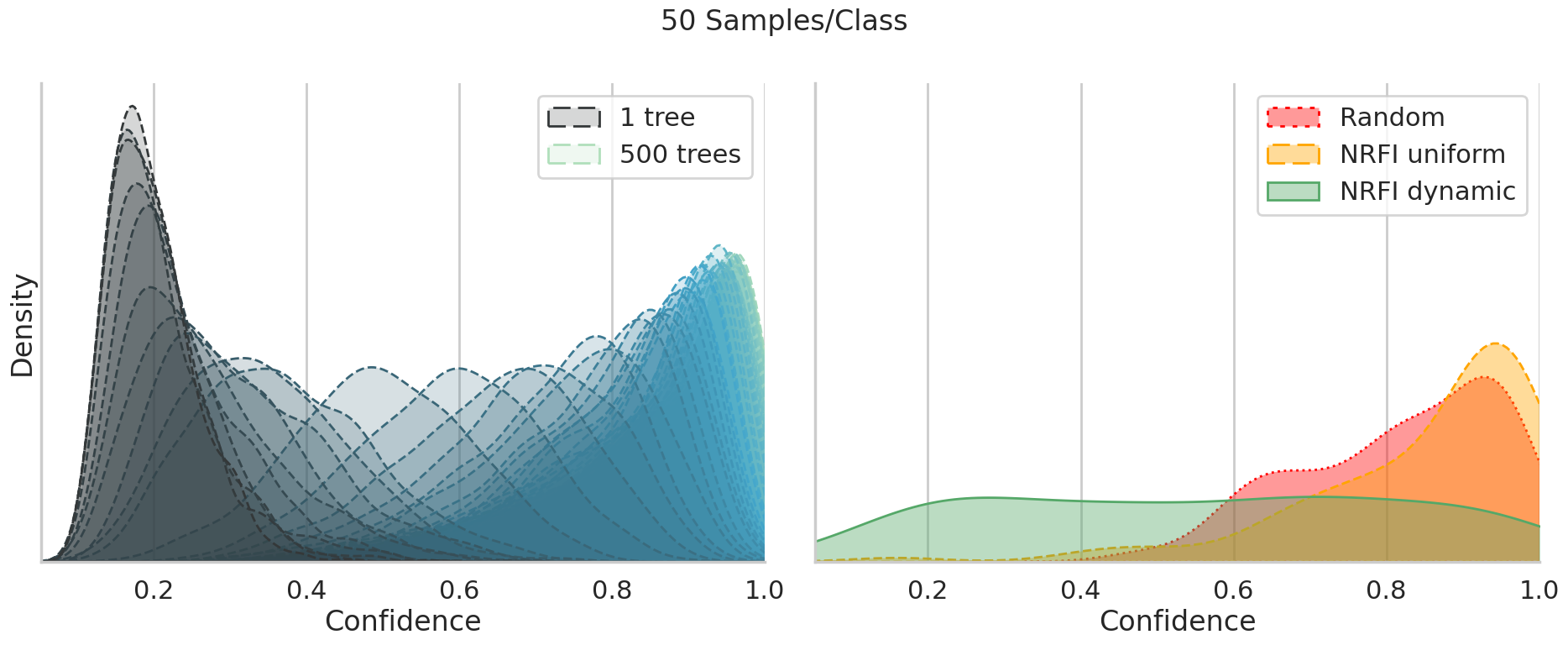}
    \end{subfigure}
	\caption{
		Probability distribution of the predicted confidences for different data generation settings on \textit{Soybean} with $5$ (top) and $50$ samples per class (bottom). Generating data with different numbers of decision trees is visualized in the left column. Additionally, a comparison between random sampling (red), NRFI uniform (orange), and NRFI dynamic (green) is shown in the right column. 
		By optimizing the decision tree sampling, NRFI dynamic automatically balances the confidences and generates the most diverse and evenly distributed data.
	}
	\label{covmap_fig_ablation_study_sampling}
\end{figure*}

\begin{table*}[t] %
	\centering
	\setlength{\tabcolsep}{6pt}	
	
	\begin{tabular}{
			lS[table-format = 2.2, table-space-text-post=\std{9.99},separate-uncertainty=true]
			S[table-format = 2.2, table-space-text-post=\std{9.99},separate-uncertainty=true]
			S[table-format = 2.2, table-space-text-post=\std{9.99},separate-uncertainty=true]
			S[table-format = 2.2, table-space-text-post=\std{9.99},separate-uncertainty=true]|
			S[table-format = 2.2, table-space-text-post=\std{9.99},separate-uncertainty=true]}
		\toprule
		& \multicolumn{4}{c}{Samples per class}& \\
		Method & {5} & {10} & {20} & {50} & {mean} \\
		\midrule
		Random       &  58.70 \std{4.15} &  58.65 \std{1.34} &  64.61 \std{6.91} &   73.24 \std{0.79} &  63.80 \std{3.30} \\
		NRFI uniform &  84.27 \std{2.57} &  87.43 \std{1.76} &  88.63 \std{1.35} &  89.52 \std{1.03} &  87.46 \std{1.67} \\
		NRFI dynamic &  84.82 \std{2.75} &  88.16 \std{1.64} &  89.10 \std{1.65} &  90.49 \std{1.47} &  88.14 \std{1.88} \\
		\bottomrule
	\end{tabular}
	
	\caption{
		Imitation learning performance (in accuracy [\%]) of different data sampling modes on \textit{Soybean}. NRFI achieves better results than random data generation. When optimizing the selection of the decision trees, the performance is improved due to more diverse sampling.
	}
	\label{covmap_table_ablation_study_sampling}
\end{table*}

Finally, we will analyze the number of parameters of the generated networks in detail. The results are shown in Table~\ref{covmap_table_network_size}. 
Current state-of-the-art methods directly map random forests into neural networks. The number of parameters of the resulting network is evaluated on all datasets with different numbers of training examples. The overall performance is shown in the last column. 
Due to the stochastic process when training the random forests, the results can vary marginally. 

Sethi, Welbl (ind-full), and Welbl (joint-full) generate networks with around \num{980000} parameters on average.
Of the four variants proposed by Welbl, joint training has a slightly smaller number of parameters compared to independent training because of shared neurons in the output layer.
Network splitting proposed by \citet{covmap_massiceti2017random} maps multiple subtrees while sharing common split nodes and reduces the average number of network parameters to \num{748000}.
Using sparse network architectures additionally reduces the number of network parameters to about \num{142000}; however, this requires a special implementation for sparse matrix multiplication.
All of the methods show a drastic increase with the growing complexity of the classifiers.
Sethi, for example, generates networks with \num{374000} parameters when training with $5$ examples per class. The average number of network parameters increases to $1.9$ million when training with $50$ examples per class.

NRFI introduces imitation instead of direct mapping. In the following, a network architecture with $32$ neurons in both hidden layers is selected.
The previous analysis has shown that this architecture is capable of imitating the random forests (see Figure~\ref{covmap_fig_compare_network_size_vs_error} for details) across all datasets and different numbers of training examples.
Our method significantly reduces the number of parameters of the generated networks while reaching the same or even slightly better accuracy. 
The current best-performing methods generate networks with an average number of parameters of either \num{142000}, if sparse processing is available, or \num{748000} when using usual fully connected neural networks. 
In comparison, neural random forest imitation requires only \num{2676} parameters.
Another advantage is that the proposed method does not create a predefined architecture but enables arbitrary network architectures. 
As a result, NRFI enables the transformation of very complex classifiers into neural networks. 

\subsection{Analysis of the Generated Data}

To study the sampling process, we analyze the variability of the generated data as well as different sampling modes in the next experiment. 
Subsequently, we investigate the impact of combining original and generated data. 

\subsubsection{Confidence Distribution}
\label{covmap_sec_ablation_study_sampling}

The data generation process aims to produce a wide variety of data samples. This includes data samples that are classified with a high confidence and data samples that are classified with a low confidence to cover the full range of prediction uncertainties.
The following analyses are shown exemplarily on the \textit{Soybean} dataset. This dataset has $35$ features and $19$ classes.	
First, we analyze the generated data with a fixed number of decision trees, i.e., the number of sampled decision trees in $RF_{\text{sub}}$. The resulting confidence distributions for different numbers of decision trees are shown in the first column of Figure~\ref{covmap_fig_ablation_study_sampling}.
When adopting the data sample to only a few decision trees, the confidence of the generated samples is lower (around $0.2$ for $5$ samples per class). 
Using more decision trees for generating data samples increases the confidence on average.

NRFI uniform and NRFI dynamic sample the number of decision trees for each data point uniformly, respectively, optimized via automatic confidence distribution (see Section~\ref{covmap_sec_confidence_distribution}). The confidence distributions for both sampling modes are visualized in the second column of Figure~\ref{covmap_fig_ablation_study_sampling}. Additionally, sampling random data points without generating data from the random forest is included as a baseline. 
The analysis shows that random data samples and uniform sampling have a bias to generate data samples that are classified with high confidence. 
NRFI dynamic automatically balances the number of decision trees and archives an evenly distributed data distribution, i.e., generates the most diverse data samples.

In the next step, the imitation learning performance of the sampling modes is evaluated. The results are shown in Table~\ref{covmap_table_ablation_study_sampling}.
Random data generation reaches a mean accuracy of $63.80\%$ while NRFI uniform and NRFI dynamic achieve $87.46\%$ and $88.14\%$, respectively.
This shows that neural random forest imitation is able to generate significantly better data samples based on the knowledge in the random forest.
NRFI dynamic improves the performance by automatically optimizing the decision tree sampling and generating the largest variation in the data.

\subsubsection{Original and Generated Data}

\begin{figure*}[t] %
	\centering
	\includegraphics[width=.8\linewidth]{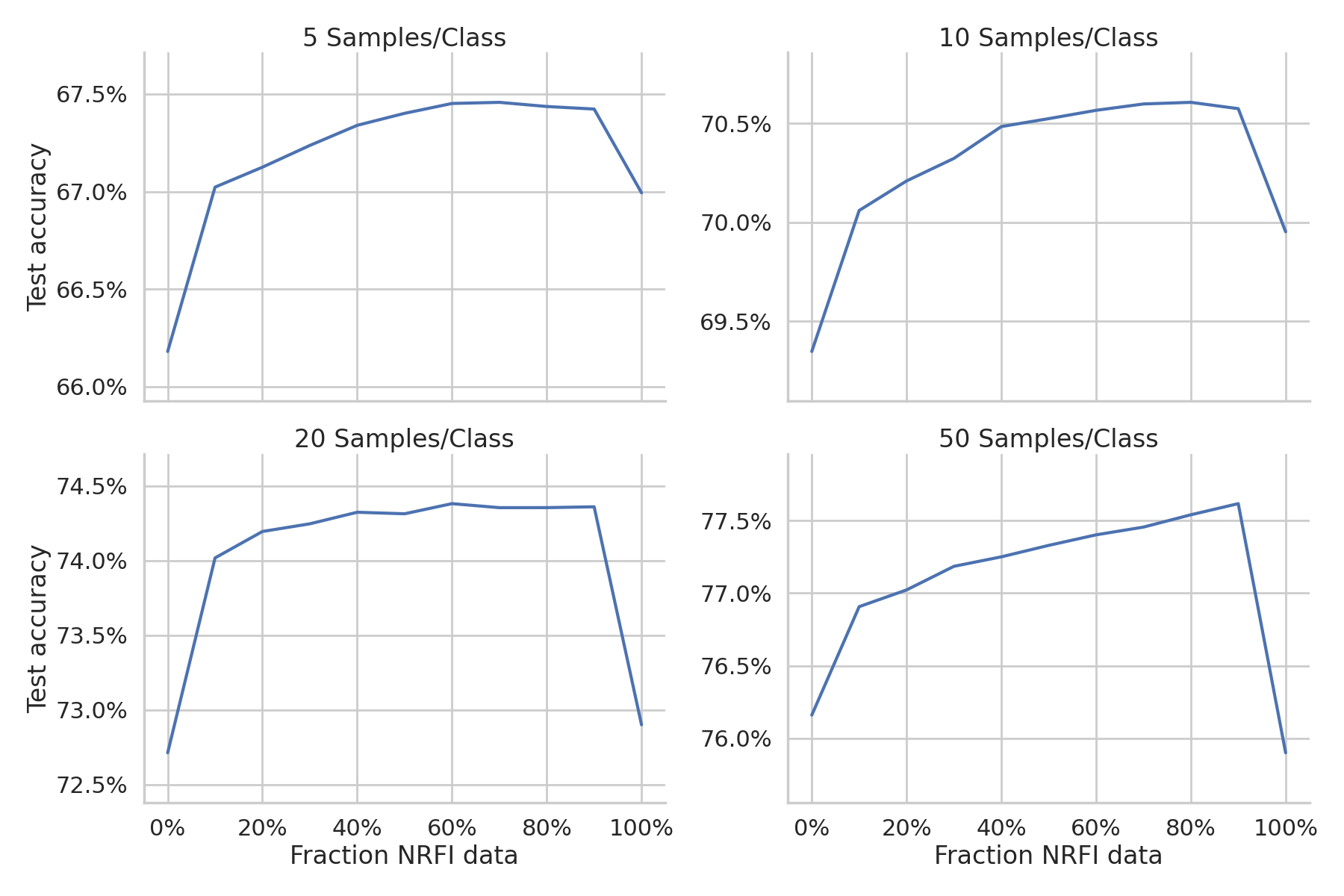}
	
	\caption{
		Analyzing the influence of training with original data, NRFI data, and combinations of both for different number of samples per class. Using only NRFI data ($w_{\text{gen}} = 100\%$) achieves better results than using only the original data ($w_{\text{gen}} = 0\%$) for less than $50$ samples per class. Combining the original data and generated data improves the performance.
	} 
	\label{covmap_fig_ablation_study_losses}
\end{figure*}

In the next experiment, we study the effects of training with original data, NRFI data, and combinations of both. For that, the 
fraction of NRFI data $w_{\text{gen}}$ is varied, which weights the loss of the generated data. Accordingly, the weight for the original data is set to $w_{\text{ori}} = 1 - w_{\text{gen}}$.
The average accuracy over all datasets for different number of samples per class is shown in Figure~\ref{covmap_fig_ablation_study_losses}. When the fraction of NRFI data is set to 0\%, the network is trained with only the original data. When the fraction is set to 100\%, the network is trained completely with the generated data. The study shows that training with NRFI data performs better than training with original data except for 50 samples per class where training with original data is slightly better. 
Combining original and NRFI data improves the performance. The best result is achieved when using mainly NRFI data with a small fraction of original data.

\section{Conclusion}

In this work, we presented a novel method for transforming random forests into neural networks. 
Instead of a direct mapping, we introduced a process for generating data from random forests by analyzing the decision boundaries and guided routing of data samples to selected leaf nodes.
Based on the generated data and corresponding labels, a network is trained that imitates the random forest.
Experiments on several real-world benchmark datasets demonstrate that NRFI is capable of learning the decision boundaries very efficiently.
Compared to state-of-the-art methods, the presented implicit transformation significantly reduces the number of parameters of the networks while achieving the same or even slightly improved accuracy due to better generalization.  
Our approach has shown that it scales very well and is able to imitate highly complex classifiers.

\section*{Acknowledgements}

This research was supported by the German Research Foundation DFG (COVMAP - RO 2497/12-2) within Priority Research Program 1894 \textit{Volunteered Geographic Information: Interpretation, Visualization and Social Computing}, the Federal Ministry of Education and Research (BMBF), Germany, under the project LeibnizKILabor (grant no. 01DD20003), the Center for Digital Innovation (ZDIN), and the German Research Foundation (DFG) under  Germany's  Excellence  Strategy  within  the  Cluster of Excellence PhoenixD (EXC 2122).

\bibliography{main}
\bibliographystyle{icml2023}

\end{document}